\documentclass[a4paper,10pt]{article}
\usepackage[margin=1in]{geometry}

\usepackage[utf8]{inputenc} 
\usepackage[T1]{fontenc}    
\usepackage{hyperref}       
\usepackage{url}            
\usepackage{booktabs}       
\usepackage{amsfonts}       
\usepackage{nicefrac}       
\usepackage{microtype}      
\usepackage{subcaption}
\usepackage{graphicx}
 \usepackage{amsmath} 
\usepackage{diagbox}          
\usepackage{authblk}

\usepackage{lmodern}

\usepackage{xspace}
\usepackage[table]{xcolor} 
\newcommand{\enb}{Easy-Options Bias\xspace}
\newcommand{\enbshort}{EOB\xspace}
\newcommand{\gda}{GroundAttack\xspace}
\newcommand{\nextqa}{NExT-QA\xspace}

\newcommand{\starqa}{STAR-QA\xspace}
\newcommand{\vmme}{Video-MME\xspace}
\newcommand{\mstar}{MMStar\xspace}

\newcommand{\rworld}{RealWorld\xspace}
\newcommand{\seed}{SEED-Bench\xspace}
\newcommand{\qwa}{Qwen2.5VL-7B\xspace}
\newcommand{\qwb}{Qwen2.5VL-3B\xspace}
\newcommand{\cpm}{MiniCPM-V2.6\xspace}
\newcommand{\vila}{ViLA-3B\xspace}
\newcommand{\seek}{DeepSeek-VL2-Tiny\xspace}

\newcommand{\darrow}{({\color{purple}$\downarrow$})}

\definecolor{CPBlue}{HTML}{003366}    
\definecolor{CPGray}{HTML}{4F4F4F}    
\definecolor{CPAmber}{HTML}{FFBF00}   
\definecolor{CPSky}{HTML}{0099CC}     
\definecolor{SPSlate}{HTML}{708090}    
\definecolor{SPMint}{HTML}{88B04B}     
\definecolor{SPCoral}{HTML}{FF6F61}    
\definecolor{SPIvory}{HTML}{FFF8E7}    

\begin{document}

\title{Mitigating Easy Option Bias in Multiple-Choice Question Answering}

\author[1,2]{Hao~Zhang}
\author[1,2]{Chen~Li}
\author[1,2,3]{Basura~Fernando}
\affil[1]{Institute of High-Performance Computing, Agency for Science, Technology and Research, Singapore}
\affil[2]{Centre for Frontier AI Research, Agency for Science, Technology and Research, Singapore}
\affil[3]{College of Computing and Data Science, Nanyang Technological University, Singapore}

%


\maketitle
\begin{abstract}


In this early study, we observe an \textbf{\enb} (\enbshort) issue in some multiple-choice Visual Question Answering (VQA) benchmarks such as MMStar, RealWorldQA, SEED-Bench, Next-QA, STAR benchmark and Video-MME. 
This bias allows vision-language models (VLMs) to select the correct answer using only the vision ($\boldsymbol{V}$) and options ($\boldsymbol{O}$) as inputs, without the need for the question ($\boldsymbol{Q}$). 
Through grounding experiments, we attribute the bias to an imbalance in visual relevance: the correct answer typically aligns more closely with the visual contents than the negative options in feature space, creating a shortcut for VLMs to infer the answer via simply vision-option similarity matching.
To fix this, we introduce \textbf{\gda}, a toolkit that automatically generates hard negative options as visually plausible as the correct answer. 
We apply it to the \nextqa and \mstar datasets, creating new \enbshort-free annotations.
On these \enbshort-free annotations, current VLMs approach to random accuracies under ($\boldsymbol{V}$+$\boldsymbol{O}$) settings, and drop to non-saturated accuracies under ($\boldsymbol{V}$+$\boldsymbol{Q}$+$\boldsymbol{O}$) settings, providing a more realistic evaluation of VLMs' QA ability.
Codes and new annotations will be released soon.

\end{abstract}


\section{Introduction}
\begin{figure}[ht!b]
  \centering
  \begin{subfigure}[t]{0.3\textwidth}
    \centering
\includegraphics[width=\textwidth]{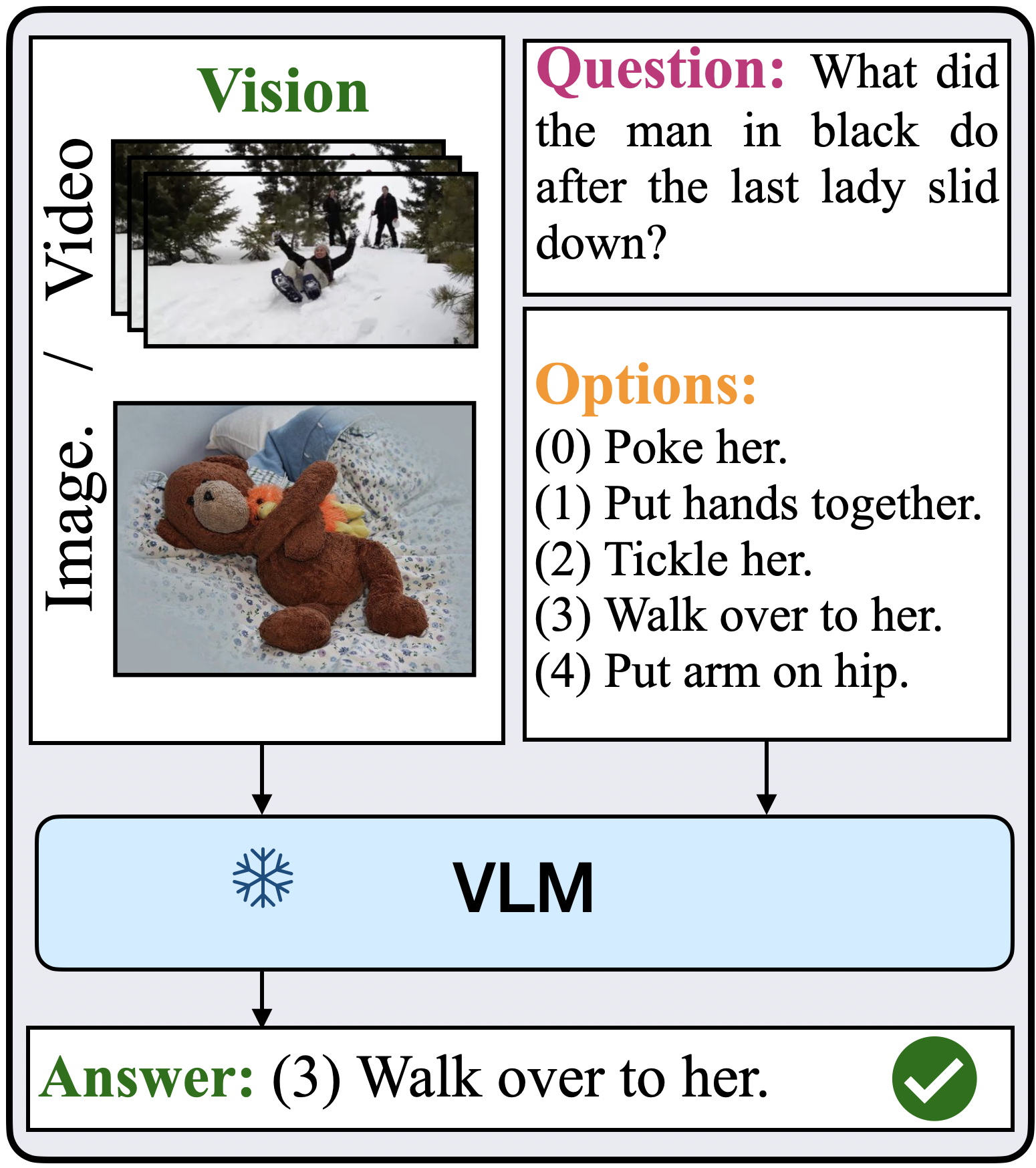}
    \caption{$\text{VLM}\left(\boldsymbol{V},\boldsymbol{Q},\boldsymbol{O}\right)\rightarrow\boldsymbol{A}$}
    \label{fig:v_q_o}
  \end{subfigure}
  \hfill
  \begin{subfigure}[t]{0.3\textwidth}
    \centering
    \includegraphics[width=\textwidth]{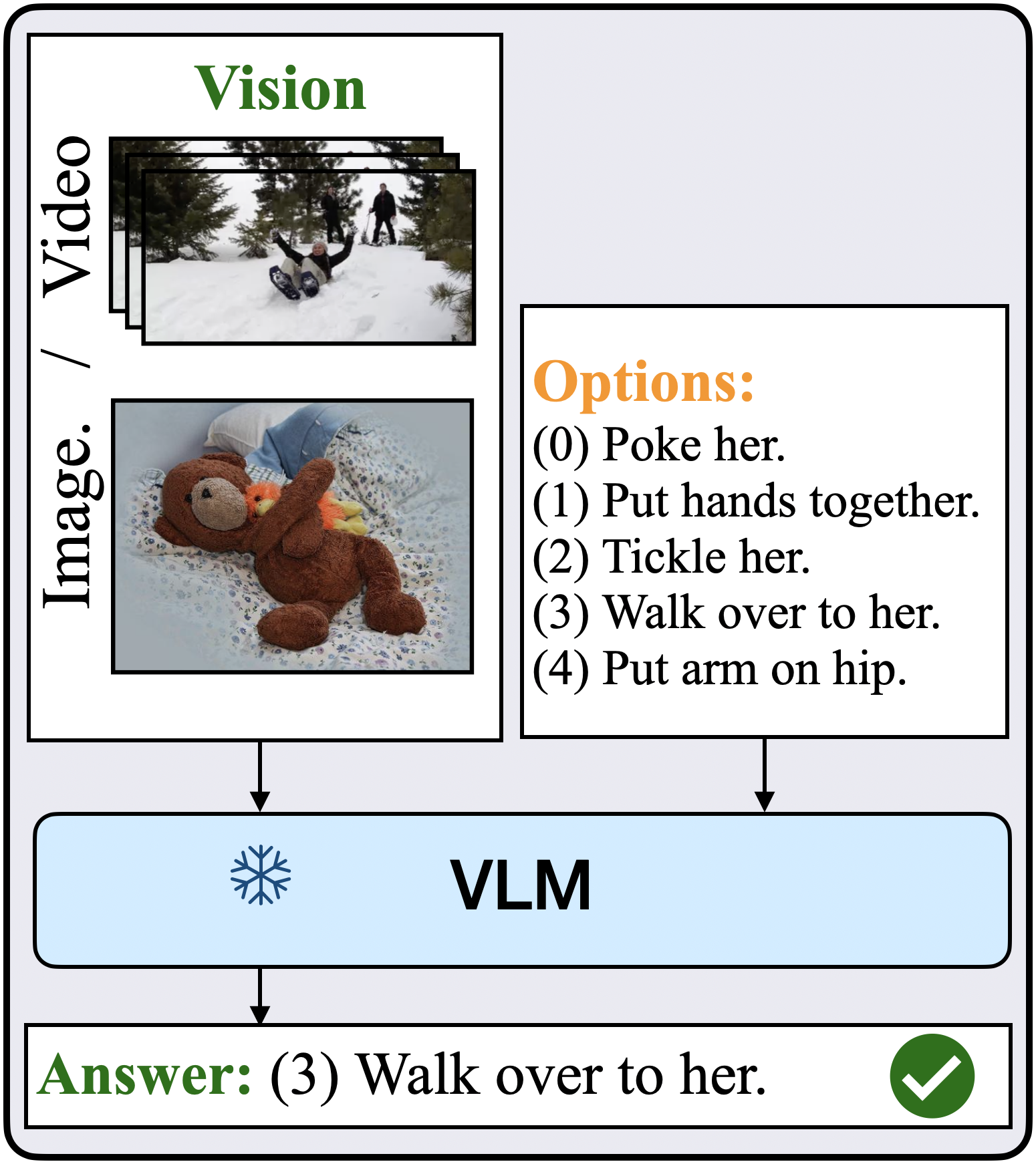}
    \caption{$\text{VLM}\left(\boldsymbol{V},\boldsymbol{O}\right)\rightarrow\boldsymbol{A}$}
    \label{fig:v_o}
  \end{subfigure}
  \hfill
  \begin{subfigure}[t]{0.3\textwidth}
    \centering
    \includegraphics[width=\textwidth]{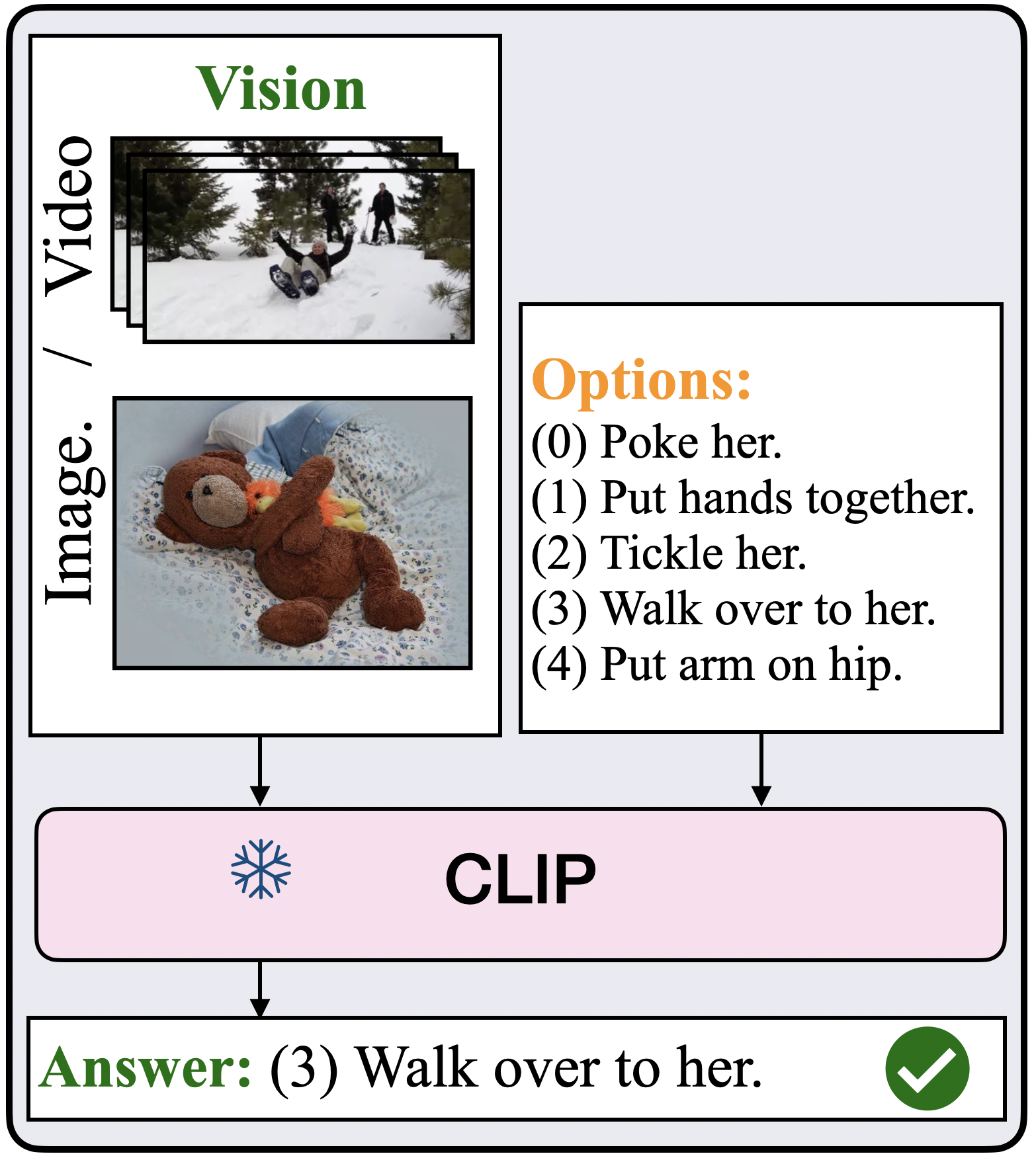}
    \caption{$\text{CLIP}\left(\boldsymbol{V},\boldsymbol{O}\right)\rightarrow\boldsymbol{A}$}
    \label{fig:v_o_clip}
  \end{subfigure}
  
\caption{\textbf{\enb} lets a VLM pick the correct answer without seeing the question. Here, $\boldsymbol{V}$, $\boldsymbol{Q}$, $\boldsymbol{O}$, $\boldsymbol{A}$ denote the vision input, question, options, and the correct answer.
}
  \label{fig:motivation}  
\end{figure}

\begin{figure}[h!tb]
  \centering
  \includegraphics[width=0.92\textwidth]{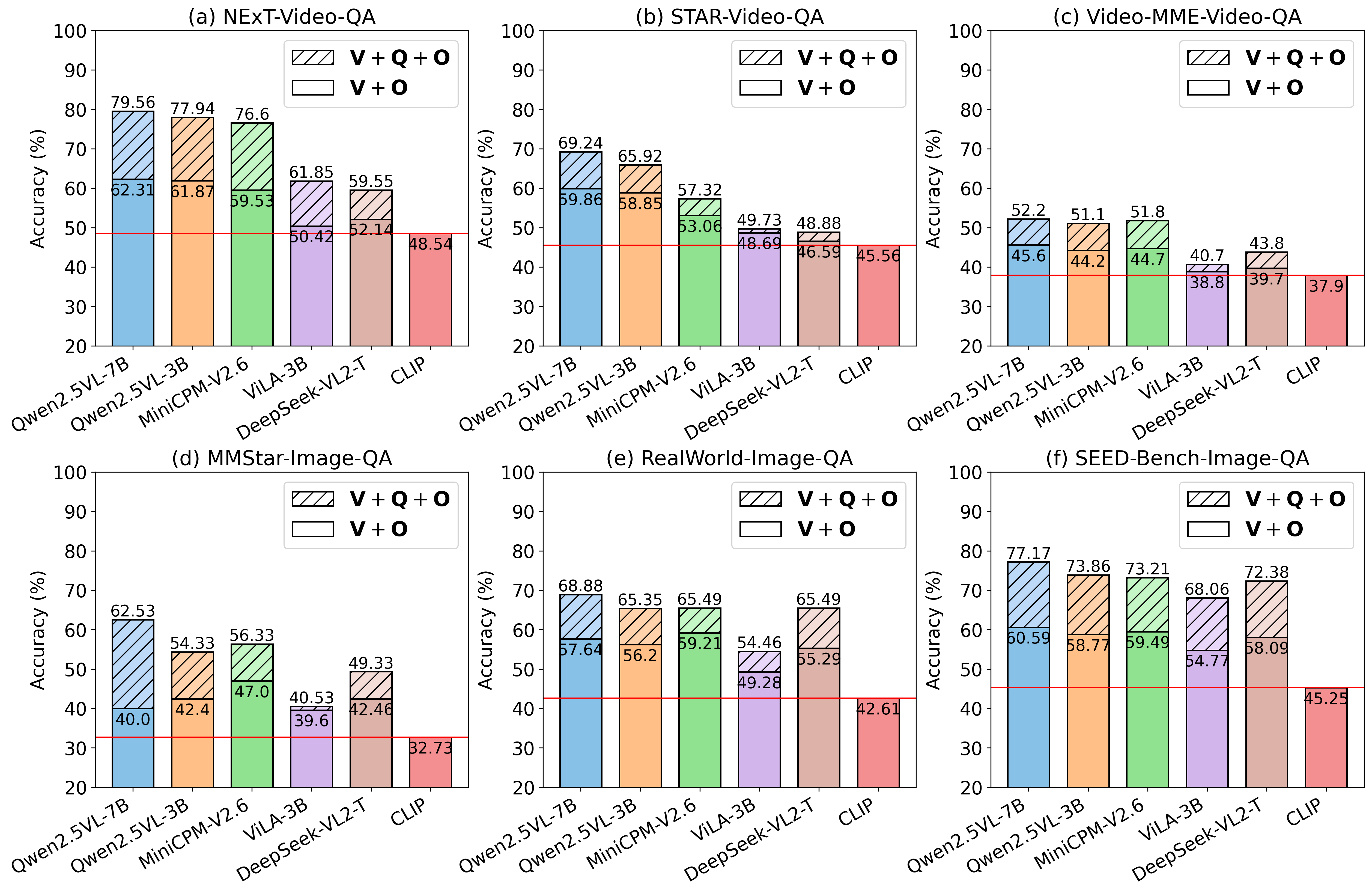}
  
\caption{%
  \textbf{\enb across six VQA benchmarks and four VLM series.} 
  Across all datasets and models, VLMs with ``vision+options'' inputs achieve an mean accuracy of 51.57\%, just 9.54\% lower than the 61.11\% mean accuracy with ``vision+question+options'' inputs. 
  When use CLIP to select the most visually similar options (``vision+options''), the mean accuracy reaches 42.1\%. 
  This shows that negative options are less groundable than the correct answer in these VQA benchmarks, creating shortcuts that VLMs can exploit.
}
\label{fig.plots.main}
\end{figure}

Visual Question Answering (VQA) is a core benchmark in multimodal research, designed to test a model’s ability to jointly reason over visual and linguistic inputs. 
In particular, multiple-choice VQA tasks require a model to select the correct answer from a set of options given an image and a natural language question. Such tasks are widely used to evaluate vision-language models (VLMs), under the assumption that correct performance necessitates understanding and integrating both the visual content and the question semantics~\cite{VQA,gurari2018vizwiz,hudson2019gqa,yu2019activitynet,xiao2021next,wu2021star_situated_reasoning,yue2023mmmu,chen2024we,lu2022learn,marino2019ok,zhang2025physreason,nguyen2025neuro,parmar2024causalchaos}.

Over the past few years, VLMs have made remarkable progress across VQA benchmarks, often surpassing human-level performance. These gains have been enabled by advances in large-scale pretraining, attention-based architectures, and integration of vision and language modalities via contrastive and generative learning. However, there is growing concern that such improvements may not reflect genuine multimodal reasoning. Instead, models may be exploiting dataset biases~\cite{pmlr-v235-rawal24a}, superficial correlations, or artifacts in the benchmark design, echoing earlier concerns in natural language processing, where models often succeed through spurious cues rather than deep understanding~\cite{geirhos2020shortcut,chao2017being,yang2020gives,manjunatha2019explicit,cadene2019rubi,clark2019don,zhong2022video}.

Before the VLM/LLMs era, \textbf{shortcut learning} \cite{dancette2021beyond} posed a critical challenge for Visual Question Answering (VQA). 
Shortcut learning happens when a model relies on shallow patterns in the <vision, question> inputs instead of truly understanding or reasoning about the contents. 
For example, on some VQA benchmarks, ``what colour...'' questions often get the answers ``white''. 
These biases emerge because the answer distributions for colour-relevant questions are long-tailed, with ``white'' appearing most frequently. 
During training, VQA models unconsciously learn these statistics. 
Since most VQA datasets split the training/testing sets in an identical distribution manner (IID), models can exploit these shortcuts to achieve spurious high accuracy on the testing set.

Several well-known biases in VQA benchmarks that lead to shortcut learning include language bias \cite{goyal2017making}, texture bias \cite{geirhos2018imagenet}, and type bias \cite{agrawal2018don}. Their key characteristic is that the model predicts the most frequent associated answer whenever a particular visual or linguistic cue appears. 
As a result, VQA models often reach performance saturation on standard benchmarks but generalize poorly to out-of-domain inputs.
To reduce shortcut learning, researchers have created de‐biased VQA benchmarks. They rebalance the testing set so it no longer mirrors the training set (OOD), forcing models to go beyond shallow patterns and combine vision and language.  
Representative works include VQA-CP \cite{agrawal2018don}, VQA-CE \cite{dancette2021beyond}, VQA-VS \cite{si-etal-2022-language}, and GQA-AUG \cite{reich2024role}. 
By breaking the training–testing correlations, these benchmarks force models to learn beyond superficial shortcuts and focus on actual understanding and reasoning.

In the VLM/LLM era, VLMs benefit from pre‐training on web data, and show strong performances on existing VQA benchmarks under \textit{zero-shot} conditions.
Unlike predecessor VQA models, which were trained on publicly available training data, VLMs are trained behind the scenes on large-scale, weakly labeled web data by industrial companies. 
That means we can't guarantee their training and testing sets meet the OOD standards.
Besides, VLMs store more prior knowledge than predecessor VQA models and learns human-like answering skills; they can exploit even subtler shortcuts than before. 
For example, Chen et al. \cite{chen2024are} show that VLMs suffer from a language bias: they can predict the correct answer using only the question, without looking at the image. 
They call this a ``\textit{lack of visual dependency}'' in current VQA benchmarks. 
When asked questions like ``Which model achieves the best ImageNet accuracy?'', a model can answer ``SoftMoE'' even without any image input. 
To address this, they gathered <vision, question, answer> triplets from six VQA benchmarks and filtered triplets with such shortcuts, forming a de-biased benchmark named MMStar.
This language‐only shortcut happens because VLMs learn prior knowledge from web‐scale data, letting them guess correctly without needing the image.

In this paper, we identify a new \textbf{\enb} (\enbshort) in multiple‐choice VQA benchmarks when testing VLMs (see Figures \ref{fig:v_q_o}–\ref{fig:v_o}). \enbshort happens when VLMs think the negative options are so irrelevant to vision inputs that they no longer need the question. 
In other words, given only ``vision+options'', a VLM can pick the correct answer just as well as if it saw the ``vision+question+options''. 
To verify that this bias appears across different tasks and models, we conduct experiments on six VQA benchmarks, including three video (\nextqa \cite{xiao2021next}, \starqa \cite{wu2021star_situated_reasoning}, \vmme \cite{fu2024video}) and three image (\mstar \cite{chen2024are}, \rworld \cite{realworldqa}, \seed \cite{DBLP:conf/cvpr/LiGGWWZS24}) datasets. 
We tested four types of SOTA VLMs (\qwa, \qwb \cite{bai2025qwen2}, \cpm \cite{yao2024minicpm}, \vila \cite{lin2024vila}, and \seek \cite{wu2024deepseek}).
When given only the vision inputs and the answer choices, VLMs still score an average of 51.57\% across all datasets and models (i.e., $\text{VLM}\left(\boldsymbol{V},\boldsymbol{O}\right)$). That’s surprisingly high, just 9.54\% below the 61.11\% average when VLMs also see the question (i.e., $\text{VLM}\left(\boldsymbol{V},\boldsymbol{Q},\boldsymbol{O}\right)$) (Figure \ref{fig.plots.main}).

This phenomenon exposes a critical flaw in the design of current VQA benchmarks: if vision-language models can consistently select the correct answer without accessing the question, then benchmark accuracy can no longer be considered a reliable measure of multimodal reasoning. We hypothesize that this \enbshort arises from several compounding factors: (1) visual bias in answer options, where correct choices trivially align with the visual content; (2) question redundancy, where the question provides little additional information beyond what is implied by the image and options; (3) shortcut learning, where models exploit spurious dataset-specific correlations; and (4) language priors, where models are overly influenced by the statistical plausibility of answer choices regardless of context.

To identify the dominant driver of \enbshort, we conduct a simple grounding experiment using CLIP~\cite{radford2021learning, zhai2023sigmoid}, a pretrained vision-language alignment model. Specifically, we compute the similarity between the visual embedding of the image (or video frames) and the text embeddings of the answer options, without including the question. Across all evaluated benchmarks, we find a striking pattern: the correct answer consistently exhibits higher visual-text alignment scores than the distractors (see Figure~\ref{fig.plots.main}). This reveals a pronounced visual relevance imbalance, where the correct answer is not only semantically appropriate but also more visually grounded than competing options.
As shown in Figure~\ref{fig:v_o_clip} and detailed in Figure~\ref{fig.plots.main}, this imbalance persists across datasets and modalities, and is sufficiently strong that CLIP alone can often identify the correct answer without ever seeing the question. This exposes a shortcut that current VLMs can exploit: by simply matching answer choices to visual features, they can bypass the question altogether, thus undermining the very objective of VQA as a multimodal reasoning task.



Addressing \enbshort is theoretically challenging due to inherent limitations in the design of the dataset. Instead, we propose \textbf{\gda}, a practical mitigation strategy that generates hard negative answer choices that are visually and semantically plausible, to rebalance option relevance. Experiments show that \textbf{\gda} significantly reduces \enbshort’s impact across benchmarks, improving the robustness of VQA evaluation. Our findings urge reconsideration of how multimodal reasoning is assessed, emphasizing the need for benchmarks where questions are indispensable.





\section{GroundAttack: creating groundable adversarial negative options}
\label{sec:gda}
\begin{figure}[h!tb]
  \centering
  \begin{subfigure}[t]{0.5\textwidth}
    \centering
\includegraphics[width=\textwidth]{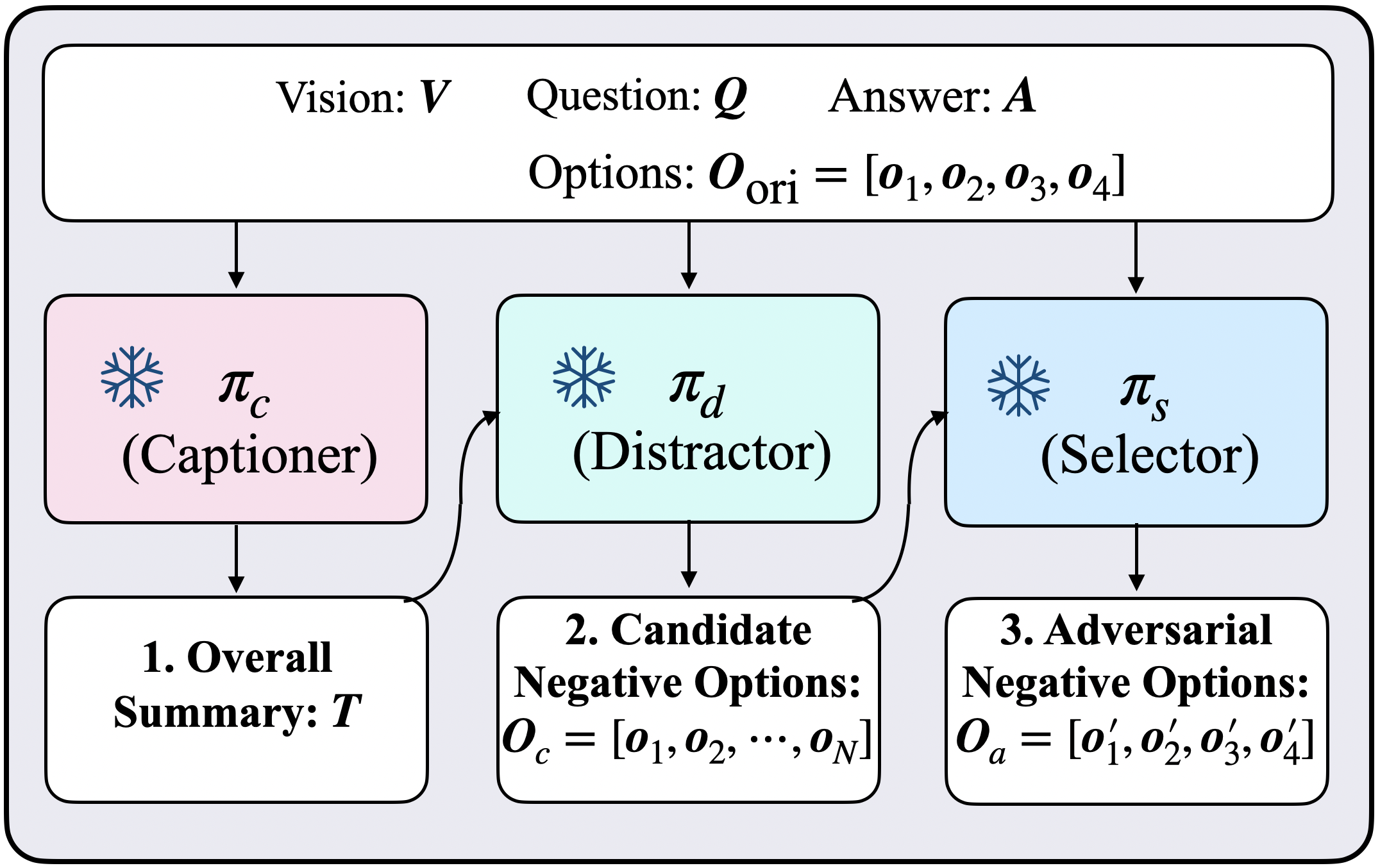}
    \label{fig:vis_question_option}
  \end{subfigure}
  
\caption{
\textbf{\gda} generates adversarial negative options that are more confusing, diverse, and visually groundable than original negatives.
It mitigates \enb in VQA benchmarks through three components:
(1) the Captioner ($\pi_c$), which converts visual content into detailed descriptions;
(2) the Distractor ($\pi_d$), which produces plausible, groundable negative candidates; and
(3) the Selector ($\pi_s$), which identifies the most adversarial negatives.
}
\vspace{-0.5cm}
  \label{fig:groundattack}  
\end{figure}

\textbf{Definition 1:} \emph{Given a visual input $\mathbf{V}$, Question $\mathbf{Q}$, options set $\mathbf{O}$ as potential answers and the correct answer $\mathbf{A} \in \mathbf{O}$, if a model $\pi(\mathbf{V},\mathbf{O})$ predicts or select the correct option answer $\mathbf{A}$, without utilizing the Question $\mathbf{Q}$ as an input to $\pi()$, then that question $Q$ suffer from \textbf{\enb~under~$\pi$}.}

\textbf{Definition 2:} \emph{Given $\mathbf{V,Q,O,A}$ tuple, if any one of the model $\pi$  from a set of models  $\Pi$ ($\pi \in \Pi $) predicts or select the correct option answer $\mathbf{A}$, without utilizing the Question $\mathbf{Q}$ as an input to it, then that question $\mathbf{Q}$ suffer from \textbf{\enb}.}

\textbf{Definition 3:} \emph{Given $\mathbf{V,Q,O,A}$ tuple,  if every model $\pi$ in a given set of models $\Pi$ selects the correct answer option $\mathbf{A}$, without utilizing the Question $\mathbf{Q}$ as an input to them, then that question $\mathbf{Q}$ suffer from  \textbf{Total~\enb}.}

\textbf{Lemma 1:} \emph{If we assume all models in $\Pi$ predicts randomly without the question as input, the expected proportion of questions suffer from \textbf{\enb} for a given benchmark is given by $1-[(1-\frac{1}{|O|})^{|\Pi|- (\lambda-1)}]$ where $\lambda$ accounts for the number of dependent models ($ 0 \le \lambda \le |\Pi|$).}

Table~\ref{tab:enb_ratio} presents the proportion of questions exhibiting \enb\ and Total~\enb\ across a range of foundation models, including \qwa, \qwb, \cpm, \vila, \seek. 
If we assume the VLM behaviour is not random, then remarkably, over 80\% of questions across all evaluated benchmarks—including recently proposed datasets explicitly designed to assess multimodal reasoning, such as \mstar, \rworld, and \seed are affected by \enb. Even more striking is the substantial fraction of questions that exhibit Total~\enb, particularly in \seed and \nextqa, indicating that in many cases, models can reliably predict the correct answer without any access to the question.
It is also worth noting that among all the benchmarks considered, \mstar exhibits the lowest incidence of \enb. However, the issue remains significant and cannot be overlooked, underscoring the need for more robust evaluation even in benchmarks designed for advanced multimodal reasoning.

\begin{table}[b]
\centering
\caption{Ratio of questions that suffers from \enb and Total \enb ~under four types of SOTA VLMs (\qwa, \qwb \cite{bai2025qwen2}, \cpm \cite{yao2024minicpm}, \vila \cite{lin2024vila}, and \seek \cite{wu2024deepseek}).}
\resizebox{\textwidth}{!}{
\begin{tabular}{l|cccccc}
\toprule
Percentage of Biased Samples &  
\nextqa &
\starqa &
\mstar&
\rworld &
\seed \\
\midrule
\textbf{\textit{\enb}} & 84.86\%& 85.24\%&78.60\%&93.72\%&80.23\%\\
\textbf{\textit{Total \enb}} & 27.17\%& 20.56\% &13.66\%&17.64\%&35.38\%\\
\bottomrule
\end{tabular}
}
\label{tab:enb_ratio}
\end{table}

Motivated by the recent works in distractor generators such as~\cite{ccavucsouglu2024disgem,chung2020bert}, we introduce \textbf{\gda} (see Figure \ref{fig:groundattack}), which addresses the \enb in existing VQA benchmarks. 
Our \gda replaces only the negative options in each tuple, while preserving the original vision, question, and answer. 
It consists of three modules: \textit{Captioner} $\pi_c$, which converts visual inputs into concise textual descriptions $\mathbf{T}$; \textit{Distractor} $\pi_d$, which generates visually grounded candidate options $\mathbf{O_c}$; and \textit{Selector} $\pi_s$, which mines adversarial hard negatives $\mathbf{O_a}$.  
We implement these modules as collaborating agents, making GroundAttack a pragmatic approach that minimizes human effort (Equations \ref{eq:caption}-\ref{eq:select}).  

\begin{subequations}
\parbox[b]{\textwidth*15/32}{
\begin{align}
&\pi_c\left({\boldsymbol{V}}\right)\rightarrow\boldsymbol{T},\label{eq:caption}\\
&\pi_d\left(\boldsymbol{Q}, \boldsymbol{A}, \boldsymbol{T}\right)\rightarrow\boldsymbol{O}_{c}, \label{eq:distract}
\end{align}}
\parbox[b]{\textwidth*15/32}{
\begin{align}
&\pi_{s}\left({\boldsymbol{V}, \boldsymbol{O}_\text{ori}, \boldsymbol{O}_{c}}\right)\rightarrow\boldsymbol{O}_a,\label{eq:select}\\
\nonumber
\end{align}}
\end{subequations}

\paragraph{Captioner $\pi_c$} Given a video or image input $\boldsymbol{V}$, model $\pi_c$ serves to convert it into a concise text description $\boldsymbol{T}$. 
Many off-the-shelf visual captioning models are available, such as BLIP \cite{li2022blip}.
These models are either trained on domain-specific video or image datasets, and fit only a small range of captioning tasks.
We resort to large-scale pre-trained VLMs as our captioner to keep \gda (GDA) compatible with different video and image benchmarks. 
This choice removes the need to pick and customize separate captioning models, greatly simplifying implementation. 

We use prompt engineering to guide the VLM in captioning complex visuals into detailed descriptions $\boldsymbol{T}$. 
A simplified prompt is shown below; full prompts appear in the supplementary material.

\fbox{%
    \parbox{\linewidth}{%
        \small{\texttt{{\color{gray}Yor are an expert video/image captioning assistant with exceptional observational and descriptive skills. Carefully analyze the provide video frames/image and generate detailed descriptions ...:}}}
    }%
}\\

\paragraph{Distractor $\pi_d$}  
Given the visual description $\boldsymbol{T}$, the question $\boldsymbol{Q}$, and the correct answer $\boldsymbol{A}$, the agent $\pi_d$ creates $N$ negative choices:$
\boldsymbol{O}_c = [\boldsymbol{o}_1, \boldsymbol{o}_2, \ldots, \boldsymbol{o}_N].
$
In past VQA benchmarks, researchers did this by hand: they either wrote rules by hand \cite{wu2021star_situated_reasoning} or hired annotators to check wrong answer options \cite{xiao2021next}. This manual process was slow but needed, since each wrong choice must be clearly wrong yet still confusing.

We now use a modern LLM to generate negative choices automatically with simple prompts. 
This change saves human effort and prevents inconsistent standards from different annotators, since the same LLM produces every negative option.  
Below is a short version of our prompt; the full prompt is in the supplementary material.

\fbox{%
    \parbox{\linewidth}{%
        \small{\texttt{{\color{gray} You are an expert in generating challenging distractors for video-based questions, Given a video description, a question and its correct answer, create 128 confusig yet incorrect answer options ... {\color{CPBlue}\{Question\}}, {\color{SPCoral}\{Correct Answer\}}, {\color{SPMint}\{Description\}}, Negative Options:}}}
    }%
}\\

\paragraph{Selector $\pi_s$}  
Given the candidate negatives $\boldsymbol{O}_c$ and the original negatives $\boldsymbol{O}_\text{ori}$, the agent $\pi_s$ chooses a subset $\boldsymbol{O}_a$ that both confuses the model and keeps enough variety among the options. 
We implement three simple strategies: \textit{random sampling}, \textit{CLIP selector}, and \textit{clustering + CLIP}. 
We experimentally verify that ``clustering + CLIP'' introduces enough confusion of negative options while maximizing options' diversity. 
Below, describe the three $\pi_s$.

\textit{Random sampling} randomly selects a fixed number of negative options from $\boldsymbol{O}_c$.  This simple baseline does not directly address the \enb, relying solely on the prior agent $\pi_d$ to ensure the negative options are confusing.

\textit{CLIP selector} picks hard negative options based on their visual similarity to the input $\boldsymbol{V}$. 
First, it extracts text features $\boldsymbol{f}_c$ for all $N$ candidates options in $\boldsymbol{O}_c$ using the CLIP text encoder (Eq.\,\ref{eq:text_feat}). 
Then, it extracts visual features for vision input $\boldsymbol{V}$ with the CLIP vision encoder; for video inputs, it applies temporal average pooling $\mathcal{F}_\mathrm{tavg}$  to compress features of multiple frames into one (Eq.\,\ref{eq:vis_feat}). 
On the contrary, for images, no pooling is needed. 
Next, it computes the similarity between each text feature and the visual feature and sorts the candidates according to similarity scores. 
Finally, it picks the top-$m$ ranked negatives with the highest similarity to $\boldsymbol{V}$ as hard negative options (Eq.\,\ref{eq:argmax2}).
. 
\begin{subequations}
\begin{align}
&\boldsymbol{f}_{c} = \mathcal{F}_{\mathrm{CLIP\text{-}Text}}(\boldsymbol{O}_c)\,,\label{eq:text_feat}\\
&\boldsymbol{f}_{v} = \mathcal{F}_{\text{tavg}}\bigl(\mathcal{F}_{\mathrm{CLIP\text{-}Vision}}(\boldsymbol{V})\bigr)\,, \label{eq:vis_feat}\\
&\boldsymbol{O}_a = \{\boldsymbol{o}_{i^*}\}_{i=1,2,\cdots, m}, \quad i^* = \operatorname*{arg\,max}_{i}\bigl[(\boldsymbol{f}_c\,\boldsymbol{f}_v^\top)_i\bigr], 
 \label{eq:argmax2}\\
&\boldsymbol{f}_c \in \mathbb{R}^{N\times D}, \quad
\boldsymbol{f}_v \in \mathbb{R}^{1\times D}\,. \nonumber
\end{align}
\end{subequations}

\textit{Clustering+CLIP} selects adversarial negative options in two steps.  
First, the K-means algorithm clusters candidate negative options into $m$ groups based on their textual features.  
Then, for each group, we use CLIP to pick the top-1 option similar to the vision input $\boldsymbol{V}$.  
We finally collect the selected options from all groups to form the groundable negative options $\boldsymbol{O}_a$ (Eq.~\ref{eq:kmeanselect}).

\begin{subequations}
\begin{align}
&[\boldsymbol{f}_{c1}, \boldsymbol{f}_{c2}, \cdots, \boldsymbol{f}_{cm}]=\text{K-Means}\left(\boldsymbol{f}_c, m\right),
\label{eq:kmeans}
\\
&\boldsymbol{O}_a = \{\boldsymbol{o}_{i^*}\}_{j=1,2,\cdots, m}, \quad i^* = \operatorname*{arg\,max}_{1}\bigl[(\boldsymbol{f}_{cj}\,\boldsymbol{f}_v^\top)_i\bigr],
\label{eq:kmeanselect}
\end{align}
\end{subequations}

We test different strategies for $\pi_s$ in \S\ref{sec:exp} and select ``Clustering+CLIP'' as it ensures that adversarial negative options are confusing, representative, and groundable.  
To this end, we built the \gda toolkit with three frozen foundation models (i.e., VLM, LLM and CLIP) in an agent-collaborating manner.

\section{Experiments}
\label{sec:exp}
\subsection{Dataset \& Setting.}
\paragraph{\nextqa{}} \cite{xiao2021next} is a widely used VideoQA benchmark comprising 5,440 videos and 34,132/4,996 QA samples in the training and validation sets.
It covers causal, descriptive, and temporal question types.
However, its negative options are randomly sampled from similar questions and manually verified, a strategy akin to $\pi_s$ = ``random sampling'', which VLMs can easily exploit with \enb.
To address this, we use \gda to generate adversarial negative options for \nextqa and will release the GDA-annotation for future research.

\paragraph{\mstar}\cite{chen2024are} is a mixed ImageQA benchmark constructed from six existing datasets that reduce language biases.
It includes an evaluation-only set of 1,500 samples spanning six categories: coarse perception, fine-grained perception, instance reasoning, logical reasoning, science \& technology, and mathematics.
We apply \gda to generate new adversarial negative options for \mstar, and will release it for future research.

\paragraph{Settings}
We use Qwen2.5VL-7B as the captioner ($\pi_c$), DeepSeek-V3 as the distractor ($\pi_d$), and SigLIP-so400m as the selector ($\pi_s$).
Using $\pi_d$, we generate $N = 128$ candidate answers and apply \gda to select $m = 4$ adversarial negatives per benchmark.
For VideoQA, only eight frames are sampled per video.
We conduct all experiments with one NVIDIA A100 81GB GPU.
To estimate LLM costs, generating candidate negatives via the DeepSeek API required processing $1.3 \times 10^7$ tokens across the \nextqa and \mstar benchmarks, costing just 8.75 USD.

\subsection{Analysis.}

\paragraph{Comparisons of different negative options.}
We compare negative options generated by the original method, random sampling, CLIP-selector, and GroundAttack in Tables \ref{tab:next_performance} and \ref{tab:mmstar_performance}.
All experiments are evaluated under the ($\boldsymbol{V}, \boldsymbol{Q}, \boldsymbol{O}$) setting unless otherwise specified.

We observe that: (1) \gda significantly decreases accuracies across all five VLMs compared to the original negative options, when \enb is mitigated.
For example, Qwen2.5VL-7B drops from 79.56\% to 50.36\%, and DeepSeek-VL2-Tiny decreases from 59.55\% to 25.80\%, which approaches the random guessing baseline of 20\% (given 5 options: 1 positive and 4 negatives). 
A similar observation appeared on the \mstar benchmark.
(2) Using random sampling in the selector $\pi_s$ introduces minimal confusion on the \nextqa benchmark (random: 79.56\% \textit{vs.} original: 65.57\% with Qwen2.5VL-7B), and even less confusion than the original annotations on the \mstar benchmark (random: 63.33\% \textit{vs.} original: 62.53\% with Qwen2.5VL-7B).
 This suggests that random sampling fails to meet the necessary criteria that negative options are confusing.
(3) Both the CLIP-Selector and \gda (Clustering+CLIP) effectively mitigate \enb on the two benchmarks.
Compared to the CLIP-Selector, \gda selects more diverse and representative negatives and produces stronger adversarial options.
(4) We further evaluate the original and \gda-generated options under the ($\boldsymbol{V}, \boldsymbol{O}$) setting, where the question is omitted.
In this setting, VLM accuracies drop to the 20\%–30\% range, close to random guessing, compared to 50\%–40\% in the standard ($\boldsymbol{V}, \boldsymbol{Q}, \boldsymbol{O}$) setting on the \nextqa and \mstar benchmarks, respectively.

\begin{table}[ht]
\centering
\caption{Comparison of VLM performance with different negative option strategies on the \nextqa benchmark.
\darrow~indicates that lower values reflect more distracting (and thus better) negative options.}
\resizebox{\textwidth}{!}{
\begin{tabular}{l|cccccc}
\toprule
\diagbox[width=3.2cm,height=1.2cm]{\small{Negative}\\\small{Options}}{\small{VLM}}
& \qwa\darrow 
& \qwb\darrow 
& \cpm\darrow 
& \vila\darrow
& \seek\darrow
\\
\midrule
Original \cite{xiao2021next}
& 79.56
& 77.94
& 76.60           
& 61.85           
& 59.55  
\\
Random Negatives
& 65.57           
& 64.11           
& 63.55           
& 49.68           
& 38.31
\\
CLIP-Selector
& \underline{51.80} 
& \underline{53.05} 
& \underline{50.46} 
& \textbf{37.19}  
& \textbf{23.38} 

\\
\rowcolor{gray!20}\textbf{GroundAttack}   
& \textbf{50.36}  
& \textbf{49.90}  
& \textbf{48.42}  
& \underline{37.85} 
& \underline{25.80}

\\ \midrule
 Original (V,O)   
& {62.31}  
& {61.87}
& {59.53} 
& {50.42}
& {52.14} 
\\

\rowcolor{gray!20}\textbf{GroundAttack}  (V,O)   
& \textbf{25.58}  
& \textbf{29.82}  
& \textbf{28.06}  
& \textbf{28.62}  
& \textbf{18.41}  
 \\

\bottomrule

\end{tabular}
}
\vspace{-0.5cm}
\label{tab:next_performance}
\end{table}

\begin{table}[ht]
\centering
\caption{Comparison of VLM performance with different negative option strategies on the \mstar benchmark.
\darrow~indicates that lower values reflect more distracting (and thus better) negative options.}
\resizebox{\textwidth}{!}{
\begin{tabular}{l|ccccc}
\toprule
\diagbox[width=3.2cm,height=1.2cm]{\small{Negative}\\\small{Options}}{\small{VLM}}
& \qwa\darrow            
& \qwb\darrow            
& \cpm\darrow            
& \vila\darrow           
& \seek\darrow 
\\
\midrule
Original \cite{chen2024are}     
& 62.53           
& 54.33           
& 56.33           
& 40.53           
& 49.33 
\\
Random Negatives        
& 63.33           
& 58.87           
& 59.80           
& 41.07           
& 39.80    
\\
CLIP-Selector  
& \underline{52.20} 
& \underline{48.40} 
& \underline{49.27} 
& \underline{30.73} 
& \textbf{27.20}  
\\
\rowcolor{gray!20}\textbf{GroundAttack}   
& \textbf{51.80}  
& \textbf{47.60}  
& \textbf{48.27}  
& \textbf{29.87}  
& \underline{30.60} \\ \midrule
 Original (V,O)   
& {40.00}  
& {42.40}  
& {47.00}  
& {39.60}  
& {42.46} \\

\rowcolor{gray!20}\textbf{GroundAttack}  (V,O)   
& \textbf{33.13}  
& \textbf{32.47}  
& \textbf{31.07}  
& \textbf{25.40}  
& \textbf{21.67} \\

\bottomrule
\end{tabular}
}
\vspace{-0.5cm}
\label{tab:mmstar_performance}
\end{table}

\begin{figure}[h!tb]
  \centering
  \begin{subfigure}[t]{0.86\textwidth}
    \centering
\includegraphics[width=\textwidth]{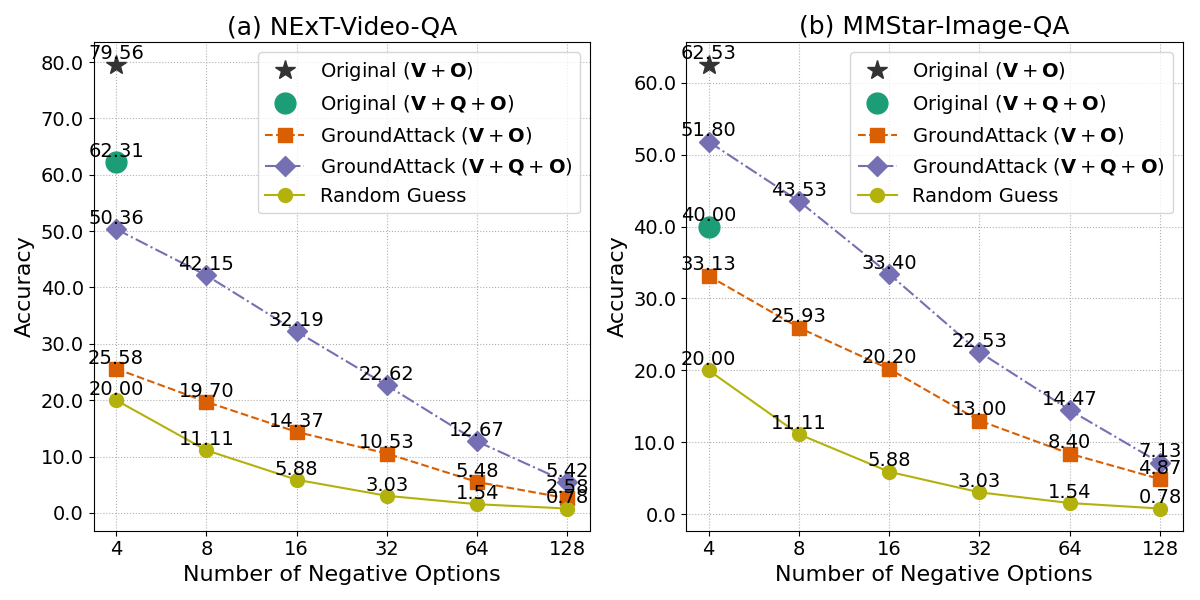}   \label{fig:accuracy_vs_num_negs}
  \end{subfigure}
\vspace{-0.5cm}
\caption{Impact of varying the number of \gda-generated negative options on \nextqa and \mstar benchmarks}
\vspace{-0.5cm}
  \label{fig:negative.size}  
\end{figure}

\paragraph{Impacts of the number of \gda negative options.}
In Figure~\ref{fig:negative.size}, we further study the impact of using different numbers of \gda-generated negative options.
 Qwen2.5VL-7B is used as the evaluation model, and tests are performed under both the ($\boldsymbol{V}, \boldsymbol{O}$) and ($\boldsymbol{V}, \boldsymbol{Q}, \boldsymbol{O}$) settings.
 As a baseline, we compute the random guess accuracy as $\frac{1}{\mathbf{O}}$ and include performance on the original annotations in the figure for reference.
We observe that accuracy decreases as the number of \gda-generated negatives increases under both settings, which aligns with expectations: more confusing distractors make it harder for VLMs to identify the correct answer.
 Notably, performance under the ($\boldsymbol{V}, \boldsymbol{Q}, \boldsymbol{O}$) setting using \gda closely approaches the random guessing curve, while original annotations do not. This suggests that \gda more effectively generates optimally confusing negative options.
Nevertheless, a noticeable gap remains between \gda and random guessing under the ($\boldsymbol{V}, \boldsymbol{O}$) setting, indicating room for further improvement in creating adversarial negative options.

\begin{table}[b]
\vspace{-0.5cm}
  \centering
  \begin{subtable}[b]{0.48\textwidth}
    \centering
    \caption{\nextqa}
    \vspace{-0.2cm}
    \resizebox{\textwidth}{!}{
      \begin{tabular}{l|cc}
        \toprule
        Percentage & \textbf{\textit{\enb}}  \darrow&  \textbf{\textit{Total \enb} \darrow}  \\
        \midrule
         Original \cite{xiao2021next} & 84.86\%&  27.17\%\\
       \textbf{\gda} & 60.36\% & \textbf{3.20}\%\\
        \bottomrule
      \end{tabular}
    }
    \label{tab:enb_ratio_nextqa}
  \end{subtable}
  \hfill
  \begin{subtable}[b]{0.48\textwidth}
    \caption{\mstar}
        \vspace{-0.2cm}
    \resizebox{\textwidth}{!}{
      \begin{tabular}{l|cc}
        \toprule
        Percentage & \textbf{\textit{\enb}} \darrow &  \textbf{\textit{Total \enb}}  \darrow\\
        \midrule
         Original \cite{xiao2021next}&78.60\%  & 13.66\% \\
       \textbf{\gda} & 67.86\% & \textbf{3.20\%}\\
        \bottomrule
      \end{tabular}
    }
    \label{tab:enb_ratio_mstar}
  \end{subtable}

  \caption{By using \gda, ratio of questions that suffer from \enb and total \enb under four types of SOTA VLMs. \darrow~indicates that less samples suffer from \enb.}
  \label{tab:enb_ratios_gda}
\end{table}

In Table~\ref{tab:enb_ratios_gda} we compute the \enb proportion after \gda on both \nextqa and \mstar datasets. Evidence suggests that our \gda enforces to models behave randomly achieving both theoretical expected performance as per the \emph{Lemma 1} for both \enb and Total~\enb.
Note that the  theoretical expected proportion for random models for both datasets is around 67\% for \enb and 3.2\% for Total~\enb.

\subsection{Visualization}
We present two examples from the original \nextqa benchmark that suffer from \enb and are corrected by \gda. Additional visualizations are provided in the supplementary material (Figure~\ref{fig:vis_example}).
\begin{figure}[h!tb]
\vspace{-0.5cm}
  \centering
  \begin{subfigure}[t]{0.98\textwidth}
    \centering
\includegraphics[width=\textwidth]{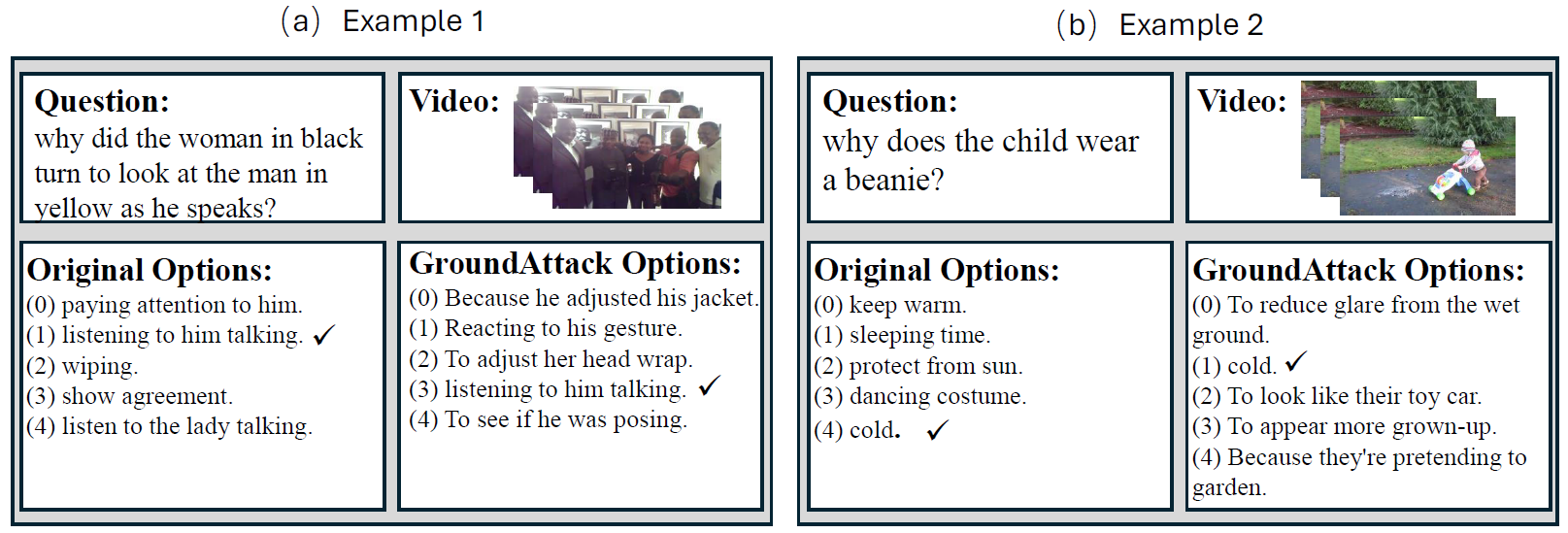}
  \end{subfigure}
\caption{Visual comparison between original negative options and those generated by \gda.}
  \label{fig:vis_example}  
\end{figure}
\section{Related works}

Several studies have uncovered critical limitations in visual question answering (VQA)~\cite{song2023recovering,qraitem2023bias,agrawal2022reassessing}, yet these findings have received limited attention in developing mainstream benchmarks.
Sheng et al.\cite{sheng2021human} demonstrated that VQA models fail dramatically when evaluated on datasets constructed using dynamic, human-adversarial approaches. Similarly, Li et al.\cite{li2021adversarial} introduced an adversarial benchmark to expose model vulnerabilities, while Zhao et al.\cite{zhao2023evaluating} examined the fragility of vision-language models (VLMs) in VQA settings. Shortcut learning in VQA, where models exploit superficial correlations across modalities, has also been addressed in several works \ cite {dancette2021beyond}.

Zhang et al.\cite{zhang2023next} identified a form of vision-answer bias in the \nextqa dataset\cite{xiao2021next}, where the distribution of answers conditioned on visual inputs is skewed, leading to disproportionate vision-answer correlations. However, their analysis was confined to conventional VQA models and did not consider recent VLMs. In this work, we reveal an even more severe issue \textbf{\enb}, where models can often ignore the question entirely and still predict the correct answer based solely on the visual input and answer choices.

Wang et al.\cite{wang2023dataset} further identified two systematic dataset biases: (1) Unbalanced Matching, where the correct answer exhibits stronger alignment with the image and question than the distractors, and (2) Distractor Similarity, where incorrect answers are both dissimilar to the correct one and mutually similar, thereby reducing discriminative challenge. A recent survey by Ma et al.\cite{ma2024robust} provides a comprehensive overview of robustness in VQA.

In contrast to prior work, we focus on how state-of-the-art VLMs, including Qwen-VL-2.5 and DeepSeek-VL2, behave under modern evaluation settings such as the \mstar, \seed, and \real  benchmarks. Our findings show the urgent need for more diagnostic evaluation protocols that account for vision-answer biases and question insensitivity in contemporary VQA tasks.
\section{Limitations and conclusions}
\label{sec:lim}
\textbf{Limitations}

Our findings are based on empirical observations and come with several limitations. First, while we analyzed the \textbf{\enbshort} phenomenon across a representative set of modern benchmarks and vision-language models (VLMs), our coverage is not exhaustive due to computational and resource constraints. As a result, the generalizability of our conclusions may be limited.

Second, although our proposed mitigation strategy shows promise in reducing \textbf{\enbshort} on the datasets and models tested, we do not claim it universally addresses the issue. Its effectiveness may vary depending on dataset properties, model architectures, and training regimes.

Third, our analysis emphasizes empirical behavior over theoretical guarantees. Understanding the root causes of \textbf{\enbshort}, such as dataset artifacts, training dynamics, or modality interplay, requires further study.

While the proposed mitigation strategy, \gda, is a practical contribution, its evaluation is limited. We have not conducted human studies to assess the quality, plausibility, or diversity of the generated distractors, relying instead on manual inspection and \enbshort reduction metrics. The grounding experiments use CLIP similarity as a proxy for visual relevance, but CLIP itself has known biases. Moreover, we did not compare our approach with alternative distractor generation methods (e.g., adversarial~\cite{li2020closer}, contrastive~\cite{cao2025enhancing,qu2024unsupervised,chung2020bert}, or perturbation-based techniques~\cite{ccavucsouglu2024disgem,geva2022break}), nor did we test whether models trained on \gda-augmented data generalize better or improve robustness across tasks.
We leave these directions for future work.

Despite these limitations, we believe our work offers valuable insights and a practical diagnostic framework for identifying and addressing a previously underappreciated failure mode in VQA tasks. We hope this encourages the development of more robust and trustworthy benchmarks for evaluating multimodal reasoning in future vision-language systems.

\textbf{Conclusions}

In this paper, we uncover a previously overlooked limitation in multiple-choice Visual Question Answering (VQA) benchmarks, which we term the Easy Negative Bias (\enbshort). This bias allows vision-language models (VLMs) to correctly answer questions without ever reading them—simply by matching visual content with answer options. Our systematic evaluation across six diverse VQA benchmarks and four state-of-the-art VLMs reveals that \enbshort is pervasive, affecting more than 80\% of questions and fundamentally undermining the credibility of benchmark accuracy as a measure of true multimodal reasoning.

To analyze the roots of \enbshort, we leverage CLIP-based grounding experiments and show that correct answer choices consistently exhibit higher visual-text alignment scores than distractors, even without the question. This imbalance suggests that many benchmarks unintentionally favour answers that are visually obvious, reducing the need for genuine cross-modal reasoning.

To address this flaw, we introduce \gda, a practical and scalable method for generating visually and semantically grounded adversarial negative options. By augmenting existing benchmarks with harder, more plausible distractors, \gda significantly mitigates \enbshort, thereby restoring the need for the question and enhancing the robustness of VQA evaluation.

Our findings call for a rethinking of how VQA benchmarks are constructed and evaluated in the era of powerful pretrained VLMs. Future benchmarks must go beyond merely testing factual recall or language priors and ensure that answering truly depends on integrating both vision and language inputs, especially the question. We hope this work spurs broader efforts toward building more diagnostic, bias-resistant tools for measuring multimodal understanding.

\noindent
\textbf{Acknowledgement}
This research/project is supported by the National Research Foundation, Singapore, under its NRF Fellowship (Award\# NRF-NRFF14-2022-0001). This research is also supported by funding allocation to B.F. by the Agency for Science, Technology and Research (A*STAR) under its SERC Central Research Fund (CRF), as well as its Centre for Frontier AI Research (CFAR).

{\small
\bibliographystyle{unsrt}
\bibliography{main}
}


\newpage
\appendix

\section{\gda and selector $\pi_s$ types}

\begin{figure}[h!tb]
  \centering
  \begin{subfigure}[t]{0.42\textwidth}
    \centering
\includegraphics[width=\textwidth]{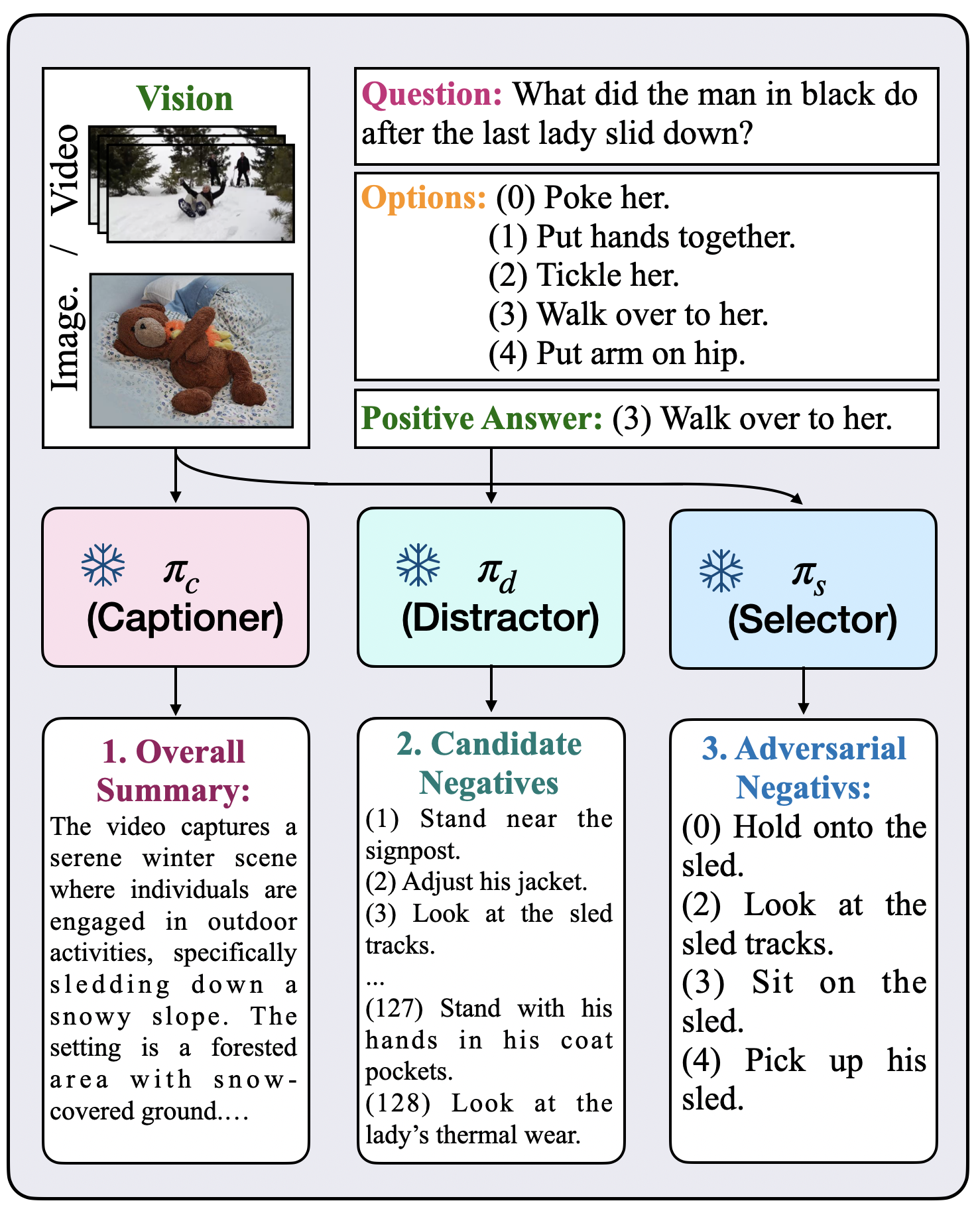}
    \caption{\gda}
    \label{fig:sub_fig1}
  \end{subfigure}
  \hfill
  \begin{subfigure}[t]{0.55\textwidth}
    \centering
    \includegraphics[width=\textwidth]{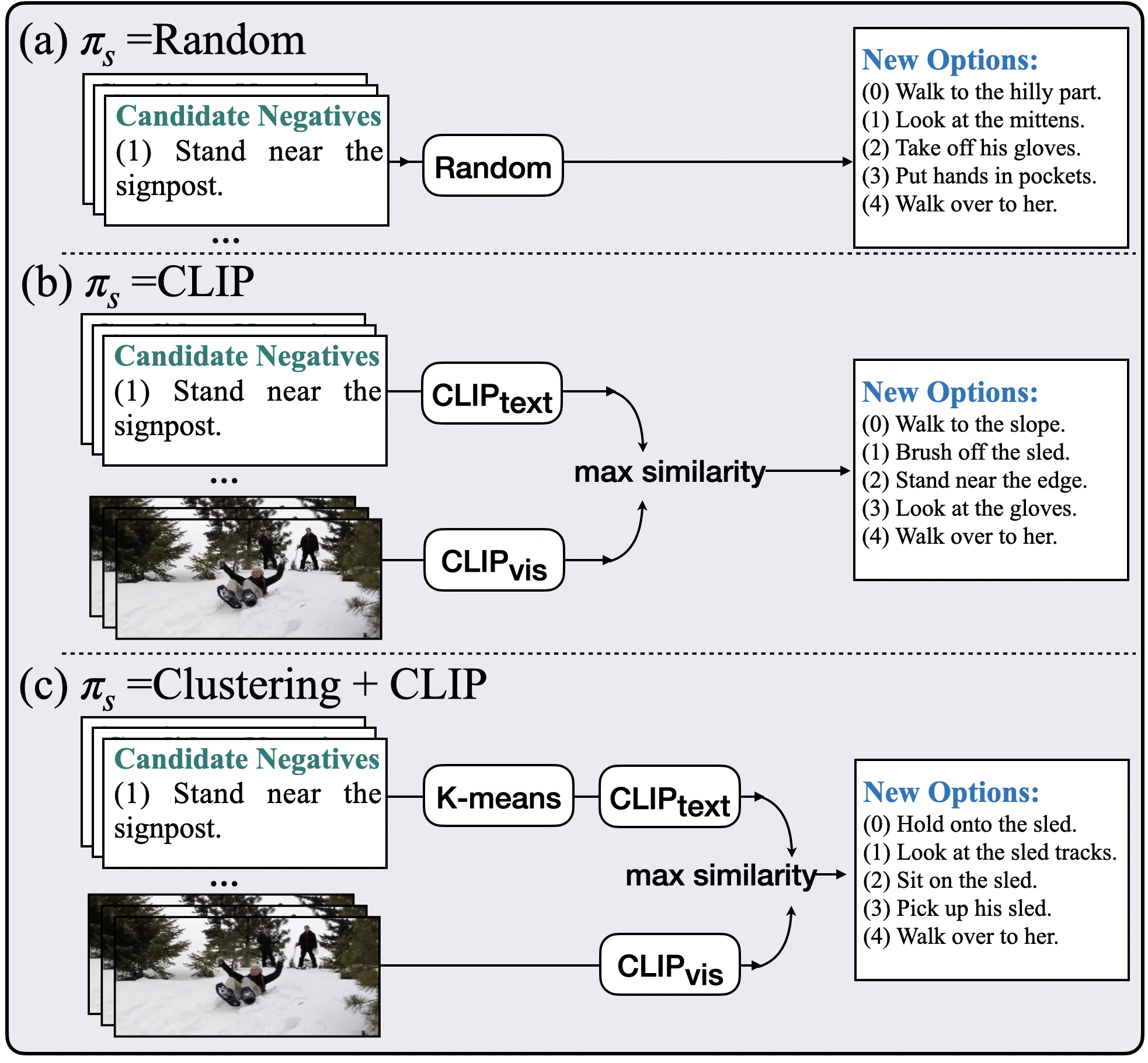}
    \caption{Selector $\pi_{s}$}
    \label{fig:sub_fig2}
  \end{subfigure}
  
\caption{
\textbf{\gda} generates adversarial negative options that are more confusing, diverse, and visually groundable than original negatives.
It mitigates \enb in VQA benchmarks through three components:
(1) the Captioner ($\pi_c$), which converts visual content into detailed descriptions;
(2) the Distractor ($\pi_d$), which produces plausible, groundable negative candidates; and
(3) the Selector ($\pi_s$), which identifies the most adversarial negatives.
}
\vspace{-0.5cm}
  \label{fig:sub_framework}  
\end{figure}

To facilitate better understanding and implementation, we present a detailed pipeline of \gda and various strategies for the selector $\pi_s$ in Figure \ref{fig:sub_framework}.
The pipeline below displays the illustrative frameworks introduced in Section~\ref{sec:gda}.

\section{Prompts for captioner $\pi_c$ and distractor $\pi_d$}

We define the roles of the captioner $\pi_c$ and distractor generator $\pi_d$ as follows:
\begin{itemize}
\item We utilize \qwa as the captioner $\pi_c$ to convert video or image-based visual inputs $\boldsymbol{V}$ into descriptive textual captions $\boldsymbol{T}$.
\item For the distractor $\pi_d$, we employ DeepSeek-V3 to generate candidate negative answers $\boldsymbol{O}_c$ conditioned on the question $\boldsymbol{Q}$, the correct answer $\boldsymbol{A}$, and the visual captions $\boldsymbol{T}$.
\item For image inputs, captions are generated based on salient objects, attributes, spatial relationships, and the overall scene.
\item For video inputs, captions focus on objects, locations, atmosphere, and dynamic actions.
\end{itemize}

We present the prompt used in \qwa for \textbf{image captions} as follows:

\fbox{%
    \parbox{\linewidth}{%
        \small{\texttt{{\color{gray}You are an expert image captioner. Analyze the image and produce a description that includes:\\
{\color{gray}1.objects}: A comprehensive list of at least 6 distinct items seen.\\
{\color{gray}2.attributes}: For each object, list its color, size, and texture.\\
{\color{gray}3.actions}: Any motions or interactions happening.\\
{\color{gray}4.spatial\_relations}: For each key pair, describe relative positions (e.g., ``cup on table'').\\
{\color{gray}5.scene}: A single sentence summarizing the setting (e.g., ``A cozy cafe interior at dusk'').\\
Use vivid, precise language and output a description.}}}
    }%
}\\

We present the prompt used in \qwa for \textbf{video captions} as follows:

\fbox{%
    \parbox{\linewidth}{%
        \small{\texttt{{\color{gray}You are an expert video captioning assistant with exceptional observational and descriptive skills. Carefully analyze the provided video frames and generate captions and detailed descriptions that vividly and accurately convey the content. Structure your response clearly, covering the following points:\\
{\color{gray}1.Overall Summary}: Briefly summarize the central theme, context, and primary setting of the video in a concise and engaging manner.\\
{\color{gray}2.Key Visual Details}: Clearly describe the main characters, significant objects, and notable locations shown in the video. Highlight distinct visual features that help viewers visualize the scenes vividly.\\
{\color{gray}3.Action Breakdown}: Provide a sequential, step-by-step breakdown of the key actions and events occurring in the video, ensuring clarity in the order of occurrence.\\
{\color{gray}4.Atmosphere and Mood}: Describe the emotional tone, atmosphere, and any stylistic elements present. Identify aspects such as lighting, music, pacing, and visual style that contribute to the video's overall mood.\\
{\color{gray}5.Text and Dialogue (if applicable)}: Include accurate transcriptions or summaries of important on-screen text or clearly audible dialogue, noting any critical details relevant to understanding the video’s context.\\
Ensure your response is articulate, engaging, and structured in a way that enables someone unfamiliar with the video to clearly visualize and comprehend its content.
}}}
    }%
}\\

We present the prompt used in DeepSeek-V3 for generating 128 \textbf{candidate negative options} as follows. Notably, the placeholders {\color{SPMint}{\texttt{Description}}}, {\color{CPBlue}{\texttt{Question}}}, and {\color{SPCoral}{\texttt{Correct Answer}}} are filled with each sample’s captioned description $\boldsymbol{T}$, question $\boldsymbol{Q}$, and correct answer $\boldsymbol{A}$, respectively:

\fbox{%
    \parbox{\linewidth}{%
        \small{\texttt{{\color{gray}You are an expert in generating challenging distractors for video-based questions. Given a video description, a question, and its correct answer, create 128 confusing yet incorrect answer options. Follow these guidelines strictly:\\
1. Grounded in the video: Each negative option must accurately describe events, actions, or details actually depicted in the video.\\
2. Specifically Incorrect: Ensure each negative option does not correctly answer the provided question.\\
3. Plausibly Confusing: Craft options that could appear correct to a model lacking deeper reasoning about temporal order, causality, object references, intentions, or relationships.\\
4. Deceptively Similar: Structure options to closely resemble the correct answer in terms of content, sequence, or entities involved, challenging superficial matching strategies.\\
5. Relevance: Avoid introducing details or actions not present in the video or clearly irrelevant to the scenario described.\\
An Example like this:\newline
[Question]: what does the white dog do after going to the cushion?\newline
[Correct Answer]: Smell the black dog\newline
[Negative Options]:\newline
(0) Lie down on the pet bed.\newline
(1) Walk towards the black dog.\newline
(2) Explore the pet bed.\newline
(3) Watch the black dog.\newline\newline
[Description]: ~~~~{\color{SPMint}\{Description\}} \newline
[Question]: ~~~~~~~{\color{CPBlue}\{Question\}}\newline
[Correct Answer]:~ {\color{SPCoral}\{Correct Answer\}} \newline
[Negative Options]:
}}}
    }%
}\\

\section{Visualization of options and \gda-generated options}
\begin{figure}[h!tb]
\vspace{-0.5cm}
  \centering
  \begin{subfigure}[t]{0.98\textwidth}
    \centering
    \includegraphics[width=\textwidth]{figure/examples.PNG}
  \end{subfigure}
\begin{subfigure}[t]{0.98\textwidth}
    \centering
    \includegraphics[width=\textwidth]{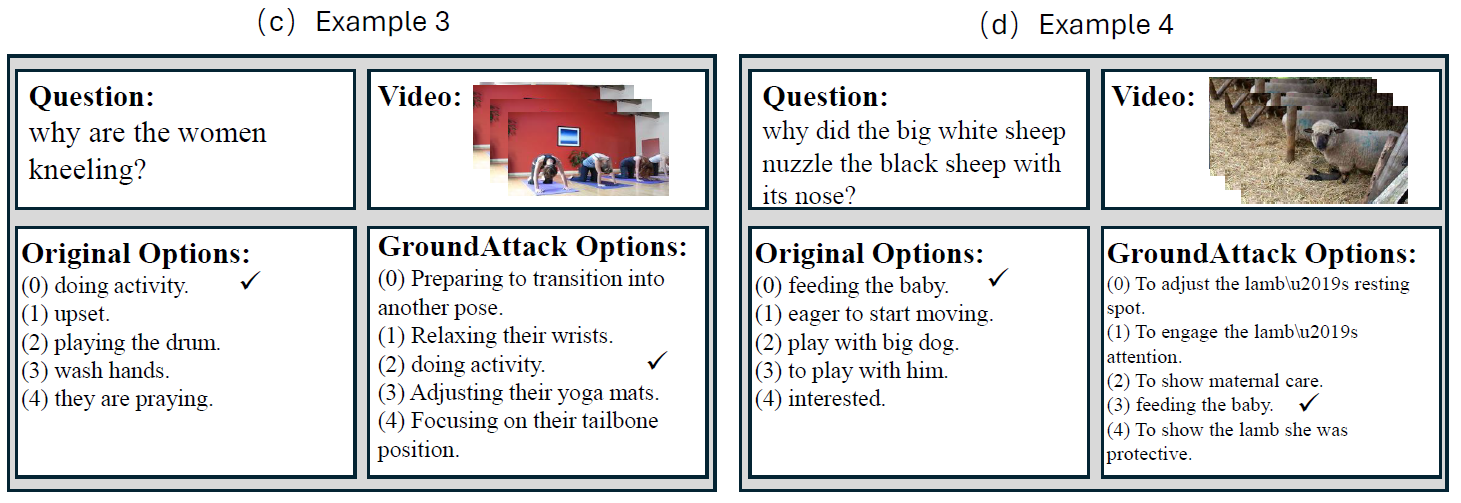}
\end{subfigure}  
\begin{subfigure}[t]{0.98\textwidth}
    \centering
    \includegraphics[width=\textwidth]{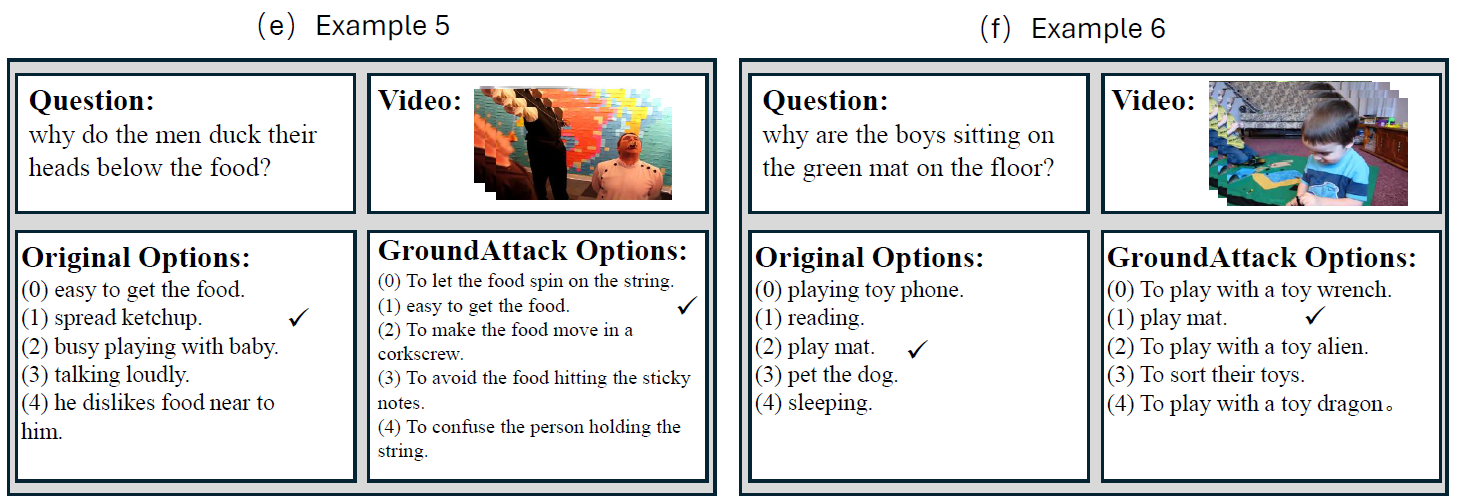}
\end{subfigure} 
\begin{subfigure}[t]{0.98\textwidth}
    \centering
    \includegraphics[width=\textwidth]{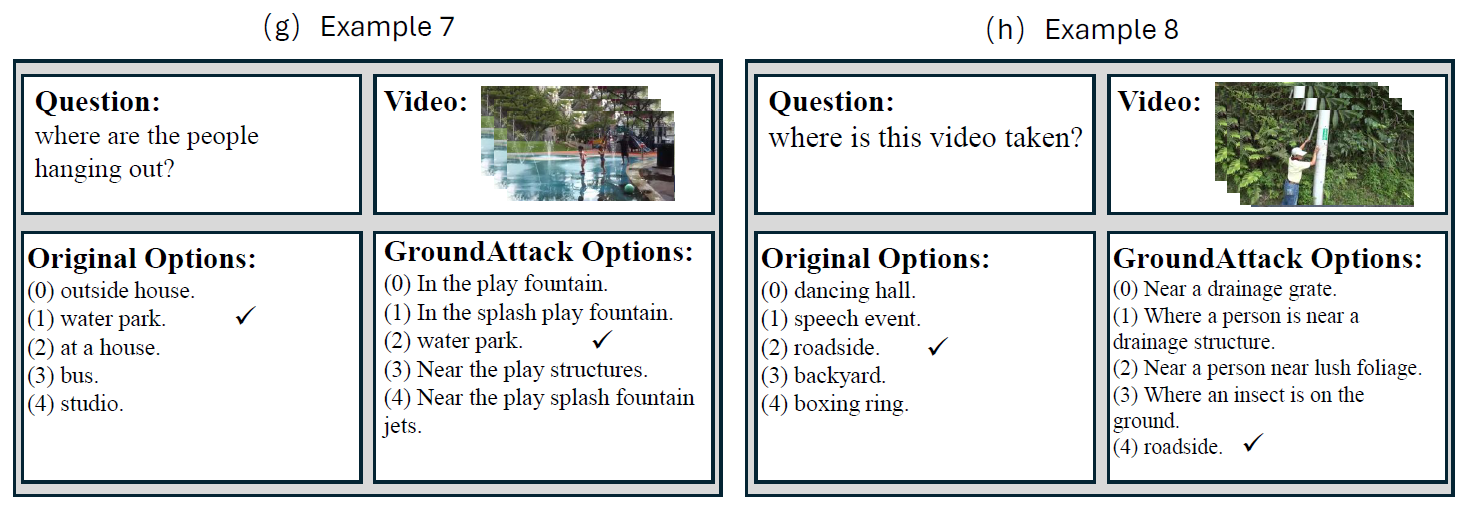}
\end{subfigure} 
\caption{Visual comparison between original negative options and those generated by \gda.}
\vspace{-0.5cm}
  \label{fig:sub_all_example}  
\end{figure}

We present additional examples from the original \nextqa benchmark that exhibit the \enb issue and are successfully corrected by \gda, as shown in Figure~\ref{fig:sub_all_example}. 
Compared to the original negative options, the \gda-generated options are more semantically grounded in the visual input, tend to be longer, and often match the length of the correct answer. 
This makes them more confusing and, thus, more challenging than the original annotations.

\section{Comparison of VLM performance with different negative option strategies on \nextqa and \mstar.}

We compare the per-category performance using negative options generated by four methods: the original annotation, random sampling, CLIP-based selection, and \textsc{GroundAttack}. 
The results are reported in Tables~\ref{tab:sub_next_performance} and~\ref{tab:sub_mmstar_performance}.

Using \qwa with only ``vision+option'' inputs on the \nextqa benchmark, causal and temporal question types perform close to random guessing (26.16\% and 20.78\% \textit{vs} a 20\% random baseline). 
This reveals a potential weakness in the original dataset: the lack of challenging, visually-groundable negative options for these question categories.
In contrast, for descriptive questions, which focus on perception-based understanding, our \gda method generates more difficult and contextually grounded negative options than the original annotations. 
As a result, \gda can reduce the prediction accuracy of vision-language models, indicating higher question difficulty and improved evaluation robustness. 

\begin{table}[h!tb]
\centering
\caption{Comparison of VLM performance with different negative option strategies on the \nextqa benchmark.
\darrow~indicates that lower values reflect more distracting (and thus better) negative options.}
\begin{subtable}[b]{0.48\textwidth}
\resizebox{\textwidth}{!}{
\begin{tabular}{l|cccccc}
\toprule
\diagbox[width=3.2cm,height=1.2cm]{\small{Negative}\\\small{Options}}{\small{\qwa}}
& Causal
& Temporal
& Descriptive
& Mean \darrow 
\\
\midrule
Original \cite{xiao2021next}
& 80.28 
& 75.06
& 86.49
& 79.56
\\
Random Negatives
& 67.89	
& 56.64	
& 76.32	
& 65.57
\\
CLIP-Selector
& \underline{54.35}	
& \textbf{42.62}	
& \underline{62.29}	
& \underline{51.80}

\\
\rowcolor{gray!20}\textbf{GroundAttack}   
& \textbf{51.75}	
& \underline{42.87}	
& \textbf{61.26}	
& \textbf{50.36}
\\ \midrule
 Original (V,O) 
& 62.56	
& 56.70	
& 73.10	
& 62.31  
\\

\rowcolor{gray!20}\textbf{GroundAttack}  (V,O)   
& \textbf{26.16}	
& \textbf{20.78}	
& \textbf{33.59}	
& \textbf{25.58}
 \\

\bottomrule

\end{tabular}
}
\end{subtable}
\hfill
\begin{subtable}[b]{0.48\textwidth}
\resizebox{\textwidth}{!}{
\begin{tabular}{l|cccccc}
\toprule
\diagbox[width=3.2cm,height=1.2cm]{\small{Negative}\\\small{Options}}{\small{\qwb}}
& Causal
& Temporal
& Descriptive
& Mean \darrow 
\\
\midrule
Original \cite{xiao2021next}
&78.10	
&74.19	
&85.20	
&77.94

\\
Random Negatives
&65.44	
&57.44	
&73.49	
&64.11
\\
CLIP-Selector
&\textbf{46.03}	
&\underline{57.92}	
&\textbf{51.54}	
&\underline{53.05}
\\
\rowcolor{gray!20}\textbf{GroundAttack}  
&\underline{51.75}	
&\textbf{42.74}	
&\underline{58.56}	
&\textbf{49.90} 
\\ \midrule
 Original (V,O) 
&61.18	
&57.51	
&73.23	
&61.87
\\

\rowcolor{gray!20}\textbf{GroundAttack}  (V,O)   
&\textbf{32.03}	
&\textbf{24.19}	
&\textbf{34.11}	
&\textbf{29.82}
 \\

\bottomrule

\end{tabular}
}
\end{subtable}
\\
\begin{subtable}[b]{0.48\textwidth}
\resizebox{\textwidth}{!}{
\begin{tabular}{l|cccccc}
\toprule
\diagbox[width=3.2cm,height=1.2cm]{\small{Negative}\\\small{Options}}{\small{\cpm}}
& Causal
& Temporal
& Descriptive
& Mean \darrow 
\\
\midrule
Original \cite{xiao2021next}
&77.75	
&72.46	
&81.34	
&76.60
\\
Random Negatives
&64.83	
&57.82	
&71.17	
&63.55
\\
CLIP-Selector
&\underline{52.51}	
&\underline{43.80}	
&\underline{57.40}	
&\underline{50.46}
\\
\rowcolor{gray!20}\textbf{GroundAttack}  
&\textbf{50.06}
&\textbf{41.56}	
&\textbf{57.14}
&\textbf{48.42}
\\ \midrule
 Original (V,O) 
&60.76	
&54.16	
&66.54	
&59.53
\\

\rowcolor{gray!20}\textbf{GroundAttack}  (V,O)  
&\textbf{30.84}	
&\textbf{22.89}	
&\textbf{29.47}	
&\textbf{28.06}
 \\
\bottomrule

\end{tabular}
}
\end{subtable}
\hfill
\begin{subtable}[b]{0.48\textwidth}
\resizebox{\textwidth}{!}{
\begin{tabular}{l|cccccc}
\toprule
\diagbox[width=3.2cm,height=1.2cm]{\small{Negative}\\\small{Options}}{\small{\vila}}
& Causal
& Temporal
& Descriptive
& Mean \darrow 
\\
\midrule
Original \cite{xiao2021next}
&61.80	
&56.95	
&72.20	
&61.85
\\
Random Negatives
&47.10	
&48.26	
&61.26	
&49.68
\\
CLIP-Selector
&\textbf{34.29}	
&\textbf{35.17}	
&\underline{51.09}	
&\textbf{37.19}
\\
\rowcolor{gray!20}\textbf{GroundAttack}  
&\underline{35.52}	
&\underline{36.35}	
&\textbf{48.78}
&\underline{37.85}
\\ \midrule
 Original (V,O) 
&49.79	
&46.53	
&60.62	
&50.42
\\

\rowcolor{gray!20}\textbf{GroundAttack}  (V,O) 
&\textbf{29.96}	
&\textbf{27.05}	
&\textbf{27.41}	
&\textbf{28.62} 
 \\
\bottomrule

\end{tabular}
}
\end{subtable}
\\
\begin{subtable}[b]{0.48\textwidth}
\resizebox{\textwidth}{!}{
\begin{tabular}{l|cccccc}
\toprule
\diagbox[width=3.2cm,height=1.2cm]{\small{Negative}\\\small{Options}}{\small{\seek}}
& Causal
& Temporal
& Descriptive
& Mean \darrow 
\\
\midrule
Original \cite{xiao2021next}
&59.46	
&54.34	
&70.66	
&59.55
\\
Random Negatives
&38.51	
&35.48	
&43.50	
&38.31
\\
CLIP-Selector
&\textbf{23.90}	
&\textbf{19.91}	
&\textbf{28.83}	
&\textbf{23.38}
\\
\rowcolor{gray!20}\textbf{GroundAttack}  
&\underline{25.66}	
&\underline{22.95}	
&\underline{32.18}	
&\underline{25.8}
\\ \midrule
 Original (V,O) 
&51.75	
&47.89	
&62.29	
&52.14
\\
\rowcolor{gray!20}\textbf{GroundAttack}  (V,O) 
&\textbf{19.49}	
&\textbf{16.69}	
&\textbf{18.40}	
&\textbf{18.41}
 \\
\bottomrule

\end{tabular}
}
\end{subtable}
\label{tab:sub_next_performance}
\end{table}

On the \mstar benchmark, we observe that \gda effectively generates challenging and confusing negative options for perception-based tasks, leading to a significant drop in accuracy for both Coarse Perception (CP) and Fine-Grained Perception (FP) question types.
 Moreover, \gda is capable of producing visually grounded hard negatives for reasoning categories, including instance reasoning and logical reasoning.
 Finally, we find that \gda generalizes well to more abstract categories such as Science \& Technology and Math, demonstrating its robustness across diverse question types.

\begin{table}[h!tb]
\centering
\caption{Comparison of VLM performance with different negative option strategies on the \mstar benchmark.
\darrow~indicates that lower values reflect more distracting (and thus better) negative options.}
\begin{subtable}[b]{0.48\textwidth}
\resizebox{\textwidth}{!}{
\begin{tabular}{l|ccccccc}
\toprule
\diagbox[width=3.2cm,height=1.2cm]{\small{Negative}\\\small{Options}}{\small{\qwa}}
& CP \darrow
& FP \darrow
& IR \darrow
& LR \darrow
& ST \darrow
& MA \darrow
& Mean \darrow
\\
\midrule
Original \cite{xiao2021next}
&72.40	
&60.00	
&69.60	
&67.20	
&43.60	
&62.40	
&62.53
\\
Random Negatives
&64.00	
&71.60	
&69.20	
&67.20	
&53.20	
&54.80	
&63.33
\\
CLIP-Selector
&\underline{41.60}	
&\underline{62.40}	
&\underline{55.60}	
&\textbf{54.80}	
&\underline{48.00}	
&\textbf{50.80}	
&\underline{52.20}
\\
\rowcolor{gray!20}\textbf{GroundAttack}   
&\textbf{40.80}	
&\textbf{59.20}	
&\textbf{55.20}	
&\underline{60.40}	
&\textbf{44.00}	
&\underline{51.20}	
&\textbf{51.80}
\\ \midrule
 Original (V,O) 
&64.80	
&36.00	
&51.20	
&40.40	
&14.40	
&33.20	
&40.00
 \\

\rowcolor{gray!20}\textbf{GroundAttack}  (V,O) 
&\textbf{34.40}	
&\textbf{31.60}	
&\textbf{40.80}	
&\textbf{38.00}	
&\textbf{25.20}	
&\textbf{28.80}	
&\textbf{33.13}  
 \\

\bottomrule

\end{tabular}
}
\end{subtable}
\hfill
\begin{subtable}[b]{0.48\textwidth}
\resizebox{\textwidth}{!}{
\begin{tabular}{l|ccccccc}
\toprule
\diagbox[width=3.2cm,height=1.2cm]{\small{Negative}\\\small{Options}}{\small{\qwb}}
& CP \darrow
& FP \darrow
& IR \darrow
& LR \darrow
& ST \darrow
& MA \darrow
& Mean \darrow
\\
\midrule
Original \cite{xiao2021next}
&69.60	
&52.40	
&59.20	
&53.20	
&32.00	
&59.60	
&54.33
\\
Random Negatives
&64.00	
&65.20	
&61.20	
&61.20	
&49.60	
&52.00	
&58.87
\\
CLIP-Selector
&\textbf{39.60}	
&\underline{57.20}	
&\textbf{48.00}	
&\underline{55.20}	
&\underline{40.40}	
&50.00	
&\underline{48.40}
\\
\rowcolor{gray!20}\textbf{GroundAttack}  
&\underline{41.60}	
&\textbf{52.40}	
&\underline{49.60}	
&\textbf{54.00}	
&\textbf{38.00}	
&\textbf{50.00}	
&\textbf{47.60} 
\\ \midrule
 Original (V,O) 
 &61.20	
 &36.40	
 &51.60	
 &43.60	
 &18.40	
 &43.20	
 &42.40
 \\

\rowcolor{gray!20}\textbf{GroundAttack}  (V,O) 
&\textbf{32.80}	
&\textbf{28.00}	
&\textbf{40.00}	
&\textbf{32.80}	
&\textbf{27.20}	
&\textbf{34.00}	
&\textbf{32.47}
 \\

\bottomrule

\end{tabular}
}
\end{subtable}
\vfill
\begin{subtable}[b]{0.48\textwidth}
\resizebox{\textwidth}{!}{
\begin{tabular}{l|ccccccc}
\toprule
\diagbox[width=3.2cm,height=1.2cm]{\small{Negative}\\\small{Options}}{\small{\cpm}}
& CP \darrow
& FP \darrow
& IR \darrow
& LR \darrow
& ST \darrow
& MA \darrow
& Mean \darrow
\\
\midrule
Original \cite{xiao2021next}
&63.60	
&49.20	
&67.60	
&54.00	
&48.40	
&55.20	
&56.33
\\
Random Negatives
&63.60	
&64.40	
&61.20	
&66.40	
&51.20	
&52.00	
&59.80
\\
CLIP-Selector
&\underline{40.80}	
&\underline{58.00}	
&\textbf{50.00}	
&\textbf{50.00}	
&\underline{45.20}	
&\underline{51.60}	
&\underline{49.27}
\\
\rowcolor{gray!20}\textbf{GroundAttack} 
&\textbf{37.20}	
&\textbf{54.00}	
&\underline{52.00}	
&\underline{52.40}	
&\textbf{42.80}	
&\textbf{51.20}	
&\textbf{48.27} 
\\ \midrule
 Original (V,O) 
 &56.80	
 &36.00	
 &60.40	
 &52.00	
 &33.20	
 &43.60	
 &47.00
 \\

\rowcolor{gray!20}\textbf{GroundAttack}  (V,O) 
&\textbf{32.00}	
&\textbf{26.00}	
&\textbf{37.20}	
&\textbf{33.20}	
&\textbf{25.60}	
&\textbf{32.40}	
&\textbf{31.07}
\\

\bottomrule

\end{tabular}
}
\end{subtable}
\hfill
\begin{subtable}[b]{0.48\textwidth}
\resizebox{\textwidth}{!}{
\begin{tabular}{l|ccccccc}
\toprule
\diagbox[width=3.2cm,height=1.2cm]{\small{Negative}\\\small{Options}}{\small{\vila}}
& CP \darrow
& FP \darrow
& IR \darrow
& LR \darrow
& ST \darrow
& MA \darrow
& Mean \darrow
\\
\midrule
Original \cite{xiao2021next}
&63.60	
&33.60	
&48.00	
&36.00	
&29.60	
&32.40	
&40.53
\\
Random Negatives
&56.80	
&46.80	
&48.00	
&38.80	
&28.00	
&28.00	
&41.07
\\
CLIP-Selector
&\textbf{33.60}	
&\underline{37.60}	
&\textbf{32.80}	
&\textbf{29.20}	
&\underline{27.60}	
&23.60	
&\underline{30.73}
\\
\rowcolor{gray!20}\textbf{GroundAttack} 
&\underline{36.00}	
&\textbf{32.40}	
&\underline{33.60}	
&\underline{30.40}	
&\textbf{23.20}	
&\textbf{23.60}	
&\textbf{29.87}
\\ \midrule
 Original (V,O) 
 &58.80	
 &31.20	
 &43.20	
 &41.60	
 &27.20	
 &35.60	
 &39.60
 \\

\rowcolor{gray!20}\textbf{GroundAttack}  (V,O) 
&\textbf{30.40}	
&\textbf{24.80}	
&\textbf{28.40}	
&\textbf{23.60}	
&\textbf{20.00}	
&\textbf{25.20}	
&\textbf{25.40}
\\

\bottomrule

\end{tabular}
}
\end{subtable}
\vfill
\begin{subtable}[b]{0.48\textwidth}
\resizebox{\textwidth}{!}{
\begin{tabular}{l|ccccccc}
\toprule
\diagbox[width=3.2cm,height=1.2cm]{\small{Negative}\\\small{Options}}{\small{\seek}}
& CP \darrow
& FP \darrow
& IR \darrow
& LR \darrow
& ST \darrow
& MA \darrow
& Mean \darrow
\\
\midrule
Original \cite{xiao2021next}
&69.20	
&43.60	
&61.60	
&40.00	
&42.80	
&38.80	
&49.33
\\
Random Negatives
&59.20	
&47.20	
&45.20	
&32.80	
&26.40	
&28.00	
&39.80
\\
CLIP-Selector
&\textbf{30.80}	
&\underline{38.40}	
&\textbf{28.80}	
&\textbf{23.20}	
&\textbf{20.40}	
&\underline{21.60}	
&\textbf{27.20}
\\
\rowcolor{gray!20}\textbf{GroundAttack} 
&\underline{32.00}	
&\textbf{36.40}	
&\underline{38.40}	
&\underline{30.40}	
&\underline{25.20}	
&\textbf{21.20}	
&\underline{30.60}
\\ \midrule
 Original (V,O) 
 &63.60	
 &30.40	
 &52.00	
 &39.60	
 &31.20	
 &38.00	
 &42.46
 \\

\rowcolor{gray!20}\textbf{GroundAttack}  (V,O) 
&\textbf{26.80	}
&\textbf{18.80	}
&\textbf{27.60}
&\textbf{20.00	}
&\textbf{17.20	}
&\textbf{19.60	}
&\textbf{21.67}
\\

\bottomrule

\end{tabular}
}
\end{subtable}
\label{tab:sub_mmstar_performance}
\end{table}


\end{document}